\documentclass{bmvc2k}

\title{EpipolarNVS: leveraging on Epipolar geometry for single-image Novel View Synthesis}

\addauthor{Gaëtan Landreau}{gaetan.landreau@meero.com}{1,2}
\addauthor{Mohamed Tamaazousti }{mohamed.tamaazousti@cea.fr}{2}

\addinstitution{
Meero\\
Paris, France\\
}
\addinstitution{
 Université Paris-Saclay,\\
 CEA-LIST,\\
 F-91120 Palaiseau, France
}

\runninghead{G. Landreau and M. Tamaazousti}{EpipolarNVS}

\usepackage{amsmath}
\usepackage{algorithm}
\usepackage{amsfonts}
\usepackage[noend]{algpseudocode}
\usepackage{caption}
\usepackage{graphicx}
\usepackage{booktabs}
\usepackage{svg}
\usepackage{amssymb}
\usepackage{makecell}
\usepackage{adjustbox}
\DeclareMathOperator*{\argmax}{arg\,max}

\usepackage{multirow}

\begin{document}

\maketitle

\begin{abstract}

Novel-view synthesis (NVS) can be tackled through different approaches, depending on the general setting: a single source image to a short video sequence, exact or noisy camera pose information, 3D-based information such as point clouds etc. The most challenging scenario, the one where we stand in this work, only considers a unique source image to generate a novel one from another viewpoint. However, in such a tricky situation, the latest learning-based solutions often struggle to integrate the camera viewpoint transformation. Indeed, the extrinsic information is often passed as-is, through a low-dimensional vector. It might even occur that such a camera pose, when parametrized as Euler angles, is quantized through a one-hot representation. This vanilla encoding choice prevents the learnt architecture from inferring novel views on a continuous basis (from a camera pose perspective). We claim it exists an elegant way to better encode relative camera pose, by leveraging 3D-related concepts such as the epipolar constraint. We, therefore, introduce an innovative method that encodes the viewpoint transformation as a 2D feature image. Such a camera encoding strategy gives meaningful insights to the network regarding how the camera has moved in space between the two views. By encoding the camera pose information as a finite number of coloured epipolar lines, we demonstrate through our experiments that our strategy outperforms vanilla encoding. 

\end{abstract}

%-------------------------------------------------------------------------

\section{Introduction}

Synthesise a novel and realistic image from another viewpoint based on single or multiple images and some camera pose information commonly referred to as novel view synthesis. It has tremendous applications, from a pure computer vision and graphics perspective (such as cinemagraph, video stabilisation or 3D-based virtual staging) to Augmented and Virtual Reality (AR/VR).

This problem can be addressed from different perspectives, depending on the available data: multiple source images, depth maps, 3D labels such as 3D scene point cloud, accurate or jittered camera poses etc. We considered in this work one of the most extreme cases for the novel-view synthesis issue. Our work constrains the prediction of a scene from a novel viewpoint by solely leveraging a single source image and the corresponding camera viewpoint transformation. 

Currently, efforts are oriented toward getting the most visually appealing results on novel view synthesis, mainly through NeRF-based methods ~\cite{NeRF,neus,MipNerf,MipNerf360}. However, only a few works pursue another tricky challenge in the novel-view synthesis: investigating the most efficient way to condition an NVS architecture on camera pose information. From a general perspective, single-image novel-view methods often restrict the viewpoint transformation to the extrinsic matrices. Intrinsic is thus discarded since most methods around single-image novel-view synthesis ignore physical image formation properties and do not account for rendering, epipolar geometry or homography concepts. Such camera pose conditioning task remains too weakly addressed in the current literature, which motivated us to tackle such an issue elegantly.

While one of the most straightforward solutions to do so consists in encoding the relative camera viewpoint transformation as a feature vector, we claim such a method is sub-optimal, supported by one of the latest state-of-the-art works in monocular depth prediction ~\cite{cameraMatters}. We thus propose in this work an elegant solution to encode the camera relative transformation as a 2D feature RGB image, by leveraging epipolar constraints. The new and implicitly encoded camera viewpoint transformation has a similar spatial resolution as the source image, somehow filling the dimensional gap between pose matrices and the RGB space. The contribution we propose in this paper is thus three-fold: 
\begin{itemize}
	\item A strategy to encode the camera pose transformation into an implicit feature image builds upon epipolar geometry considerations. 
	\item A neural network architecture which leverages such camera viewpoint transformation encoding. 
	\item A spectral loss function that extensively accounts for the higher frequencies of an image to better retrieve tiny and complex details.
\end{itemize}

\section{Related work}

\noindent\textbf{Novel view synthesis in a large extent.} Since modalities involved in novel-view synthesis are broad (images, video sequence, 3D point clouds, depth and disparity maps, camera poses), tremendous approaches investigated ways to tackle such issues. One of the latest trends takes advantage of Neural Radiance Fields (NeRF ~\cite{NeRF}) and their incredible powerfulness to generate highly realistic scenes from unseen viewpoints ~\cite{neus,Regnerf,MipNerf360}. However, such architectures are somehow over-fitted over a unique scene and do not have any generalisation abilities. Such a drawback is growingly overcome by recent works ~\cite{pixelnerf,mine} that tackle novel view synthesis through the prism of NeRF-based architectures. While MINE ~\cite{mine} is a Multiplane Images (MPI) based method that requires accurate ground truth disparity maps, PixelNeRF ~\cite{pixelnerf} works with several input views to refine the predicted novel view. Another important line of work in novel view synthesis for indoor navigation, with datasets such as RealEstate10K ~\cite{RE10k} or Matterport3D ~\cite{Matterport3D}, produces extremely appealing results with complex architectures ~\cite{Synsin,geoFree,PixelSynth}, often at the expense of complex information to get, such as ground truth depth maps or dense 3D point clouds. \newline

\noindent\textbf{Camera pose encoding.} Camera pose is compactly represented through: 3 degrees of rotations around each world axis, 3 other ones for translation (both defining the extrinsic matrix) and a few additional ones when intrinsic must be considered (focal length, sensor size etc). One of the most straightforward solutions encodes the viewpoint transformation by computing camera poses difference ~\cite{sun2018multiview}. Another pose encoding strategy embeds such a low-dimensional camera pose into a higher space, as in ~\cite{NVS_skip,geoFree}. These last two possibilities can be simplified if one wants to consider one-hot vectors for camera pose encoding. Finally, the closest work to ours regarding camera pose encoding is ~\cite{cameraMatters}, which encodes the camera location (parametrized through a roll and pitch angles as well as a fixed height above a ground plane) as a 2D feature image for depth maps prediction purpose. \newline

\noindent\textbf{Camera pose conditioning.} Extrinsic camera pose is thus often the unique 3D prior information that conditions the neural network for generating a novel view. Based on the camera pose difference $P_{diff}=P_{target}-P_{source}\in \mathbb{R}^{v}$, the authors from ~\cite{sun2018multiview} tiled such vector across all the pixels of the source RGB image, feeding their CNN-based architecture with inputs that size $\mathbb{R}^{H\times W\times (3+v)}$. On the other hand, ~\cite{NVS_skip} adopts a different strategy and concatenates its camera pose feature vector with the one obtained from their CNN-encoder before feeding it to the decoder. 
Finally, ~\cite{Synsin} designed a single image novel-view synthesis method that extensively relies on 3D point cloud consideration. Camera viewpoint transformation is used within the network architecture to update the predicted point cloud before rendering. \newline

\noindent\textbf{Single-image novel view synthesis.} Such a framework is the most challenging one since only minimal information is available during training and inference: a source image and a corresponding camera pose transformation that accounts for the target image we aim to generate. To the best of our knowledge, only a few recent works ~\cite{sun2018multiview,NVS_skip,pixelnerf}  deal with such a highly constrained setting. While ~\cite{sun2018multiview, NVS_skip} both handle "discrete" (from ShapeNet ~\cite{Shapenet}, parametrized through a unique azimuthal angle, the elevation one being fixed) and continuous (as in Synthia ~\cite{Synthia} and KITTI ~\cite{KITTI}) camera transformations, the pose-feature vector needs to be updated accordingly from a size perspective. This is one of the main benefits of the method we designed. Both discrete and continuous camera information is encoded through a featured image that sizes the same as the source image and that truly leverages the real-world camera transformation that occurred between the source and the target view. Such convenient property allows to inferring (at least with discrete camera poses) viewpoints that were not represented within the training set. \newline

\section{Method}
\subsection{Camera viewpoint transformation encoding}
\subsubsection{Epipolar geometry overview }

The general framing of our work might be considered one of the trickiest ones in novel-view synthesis since the image generation from a different camera viewpoint is only made prior to a single source image and a relative camera transformation. 

We denote by $I_{s} \in \mathbb{R}^{H\times W\times 3}$ the RGB source image and $I_{t}$ the target one we aim to predict.
The pinhole convenient camera model that we get consideration for is represented through an intrinsic matrix $K \in \mathbb{R}^{3\times3}$. The rigid motion that accounts for the relative transformation between the source and the target view consists of a rotation $R \in SO(3)$ and translation $T\in \mathbb{R}^{3\times1}$, expressed through each camera's extrinsic:
\begin{equation}
     \begin{cases}
     R = R_{t} R_{s}^{T} \\
     T = t_{t} - R t_{s}
     \end{cases}
\end{equation}
with $(R_{s},t_{s})$ and $(R_{t},t_{t})$ respectively accounting for the source and target camera extrinsic. Epipolar geometry ~\cite{epi1} has consideration for the projective geometry that connects two camera viewpoints and has various applications such as Structure from Motion ~\cite{tamaazousti2011nonlinear}. Epipolar geometry aims to describe the relationship that stands between 3D world location and 2D pixel coordinates, given a stereo pair of cameras and their corresponding poses. The fundamental matrix F is a $3\times3$ matrix that entirely describes such 3D/2D mapping and can be obtained through: 
\begin{equation}
    \mathbf{F} = K^{-T} [T]_{X} R  K^{-1}
\end{equation}

with $[.]_{X}$ the skew-symmetric matrix representation of any one-dimensional vector. Given a pixel location\footnote{Homogeneous coordinates are implicitly used here but omitted for clarity reason} $p_{s}\in I_{s}$, such fundamental matrix $\mathbf{F} \in \mathbb{R}^{3\times3}$ allows to define: 
\begin{equation}
    \mathcal{P}_{p_{s}} = \{p_{t}\in I_{t} | p_{t}^{T}\mathbf{F}p_{s} = 0 \}
\end{equation}

as the finite set of pixels from $I_{t}$ that live on the epipolar line defined by $l=\mathbf{F}p_{s}$. From a pure geometrical perspective, such line corresponds to the rendered (on $I_{t}$ camera plane) 3D ray that passed through both the camera center of $I_{s}$ and $p_{s}$. 

The fundamental matrix $\mathbf{F}$ makes a pixel-to-line correspondence through a linear equation that involves both the source and the target original camera location. As soon as we aim to use the epipolar geometry to encode the viewpoint transformation, a sampling strategy needs to be set in order to determine which pixel location from the source image $I_{s}$ are going to be used to compute these epipolar lines. Instead of randomly sampling location over the $H\times W$ possibilities, and motivated by experiments that can be found in the Supplementary, pixels are sampled according to a regular grid $\textbf{G}_{r}$ that spans the whole image, parameter \textit{r} controlling how coarse the grid is: 

\begin{equation}
    \mathbf{G}_{r} = \left\{(p_{x},p_{y}) \in \{1,..,H\}\times \{1,..,W\} \Big\rvert \begin{array}{l}
                    p_x \equiv 0 \pmod{H/r}\\
             p_y \equiv 0 \pmod{W/r}
              \end{array}\right\}
\end{equation}

\subsubsection{Encode the camera viewpoint transformation}

Our module encodes the relative camera motion it exists between the source and target views by extensively leveraging epipolar geometry. Its output is fed to one of the branches of our NVS neural network as a feature image, that thus implicitly represents the transformation that occurred, and is referred as $E_{s\xrightarrow{}t}$.
An overview of the different stages involved to compute $E_{s\xrightarrow{}t}$ is presented through the pseudo-code\footnote{Some pixel values on $E_{s\xrightarrow{}t}$ might be overwritten by another pixel $p$. We claim the sampling order over \textbf{G} has no impact on the encoding strategy.} below in Algorithm \ref{pseudoCode}. \newline

As seen in Algorithm \ref{pseudoCode}, each epipolar line reported on $E_{s\xrightarrow{}t}$ has a distinct colour, the one associated with the sampled pixel on $I_s$. Such implementation choice gives additional RGB prior information to the network regarding the colour that should be generated, even though lightning issues are not considered. One might notice that the last epipolar lines plotted on $E_{s\xrightarrow{}t}$ would overwrite some previous ones, at least at some specific pixel location (on epipoles for instance). We claim, based on experimental observations, that pixel sampling order (and thus epipolar line overwriting) does not have an impact that is significant on the training and inference performances of the model. 

The encoding method depicted here is thus able to encode any form of camera viewpoint transformation without any structural adaptation. Indeed, most concurrent works need to change the way viewpoint transformation is encoded since continuous poses are not processed the same way as discrete ones, where a one-hot encoding strategy might be used. 

\begin{algorithm}[h!]
\caption{Epipolar Encoding module \label{pseudoCode}}
\begin{algorithmic}[1]
\Procedure{Input: $(I_{s},F,\mathbf{G}_{r})$}{}
\State $E_{s\xrightarrow{}t} = \textit{zeros(H,W,3)}$
\For {$p_{G}$ in $\mathbf{G}_{r}$}
\State $colRGB=I_{s}[p_{G}]$
\State \text{Build up} $\mathcal{P}_{p_{G}}$
\State $\forall p \in \mathcal{P}_{p_{G}},\hspace{.2cm} E_{s\xrightarrow{}t}[p]=colRGB$
\EndFor
\State Return $E_{s\xrightarrow{}t}$
\EndProcedure
\end{algorithmic}
\end{algorithm}

\begin{figure}[h!]

\begin{center}
\includegraphics[width=.26\textwidth]{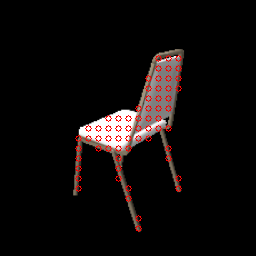}\hspace{.5cm}%\hfill
\includegraphics[width=.26\textwidth]{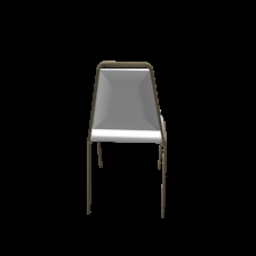}\hspace{.5cm}%\hfill
\includegraphics[width=.26\textwidth]{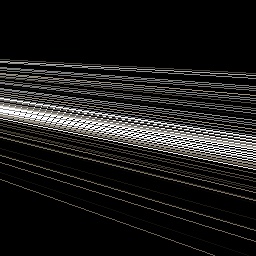}
\end{center}
\caption{From left to right: Source image $I_s$, Target image $I_t$ and the corresponding $E_{s\xrightarrow{}t}$. From the ShapeNet ~\cite{Shapenet} \textit{chair} class. Pixel locations sampled from $\textbf{G}_{25}$ are highlighted through red circles.}
\label{fig:examplePoseEncoded}
\end{figure}

Only non-null pixel values are sampled from $\textbf{G}_{r}$ in our encoding strategy for ShapeNet ~\cite{Shapenet} since pixels located in the background do not bring any valuable information regarding the corresponding coloured epipolar lines. Figure \ref{fig:examplePoseEncoded} represents the kind of results one might expect with our viewpoint camera transformation encoding strategy on ShapeNet ~\cite{Shapenet}. 

\subsubsection{Extended strategy for encoding}

Performing novel view synthesis on Synthia ~\cite{Synthia} and KITTI ~\cite{KITTI} datasets is, from a relative camera transformation perspective, somehow redundant and tricky through our encoding framework. Indeed images contained in these datasets have been recorded through a car driving across city streets, and most of the camera transformations considered are translation motions. Such viewpoint changes between the source and target views are improperly handled by the epipolar theory since depth cues are lost. We therefore extended our initial encoding strategy for those datasets through a fourth channel that primarily accounts for such depth information.

Let's denote:
\begin{equation}
    \Delta_{t}= |t_{t}| - |t_{s}| = \left[\Delta t_{X},\Delta t_{Y},\Delta t_{Z} \right]^{T} \in \mathbb{R}^3
    \label{eq:delta_t}
\end{equation}
the difference between the two absolute translations $t_s$ and $t_t$. Absolute values are taken in Equation \ref{eq:delta_t} since 2D planes coordinates are not standardised across scenes in KITTI ~\cite{KITTI} and Synthia ~\cite{Synthia}. The meaningful scalar value of interest is referred as $\delta_{t}$ and is computed through:
 
 \begin{equation}
 \label{eq:2}
     \delta_{t} = sign(t_{M}) \times| t_{M} |
 \end{equation}
 with \newline
 \begin{equation}t_{M} = \Delta_{t}\left[p_{M}\right] \hspace{.3cm} ;  \hspace{.3cm} p_{M}=\argmax |\Delta_{t}|\end{equation}

The expression of $\delta_{t}$ satisfies two resourceful constraints in our case: 

\begin{itemize}
    \item It accounts for the main car motion direction and gives some insight regarding the distance the car moved between the source and target view.
    \item The sign of $\delta_{t}$ indicates whether the source/target frames that were sampled correspond to a forward or backward motion. 
\end{itemize}
Such properties allow the network to better apprehend the direction of the motion it should take into consideration. The value of  $\delta_{t}$ is finally stored on a fourth channel of $E_{s\xrightarrow{}t}$ only at locations where the epipolar lines had non-zero values in the first three RGB channels. 

\subsection{Network architecture and Training loss}

\begin{figure}[h!]
  \begin{center}
  \includegraphics[width=.8\textwidth]{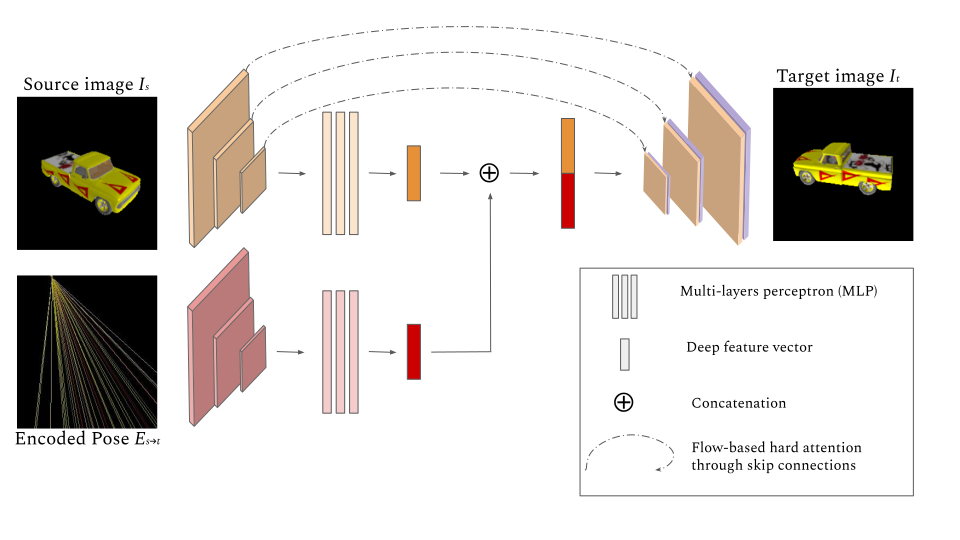}
  \end{center}
  \caption{General overview of our architecture. Network takes as inputs both $I_s$ and $E_{s\xrightarrow{}t}$ through two distinct encoders that produce feature vectors that are concatenated before being fed to the decoder. The network structure also leverage on the hard flow attention connections introduced in ~\cite{NVS_skip}.}
  \label{fig:architecture}
\end{figure}

The overall network architecture is presented in Figure \ref{fig:architecture}. Such architecture takes inspiration through the one ~\cite{NVS_skip} introduced in their work, at least regarding the image-to-image U-Net based encoder/decoder structure with the hard-flow attention strategy. However, the way the transformation viewpoint information is provided to the network architecture drastically changes for our model. While ~\cite{NVS_skip} performed feature-vectors concatenation at the network's bottleneck stage, we claim such choice is sub-optimal, at least for discrete camera pose information as contained in ShapeNet ~\cite{Shapenet}. We thus rather encode this camera pose transformation through $E_{s\xrightarrow{}t}$ as an image that feeds a second CNN-based encoder. This last encoder produces a feature vector that will be concatenated to the one obtained by the U-Net based encoder before being consumed by the decoder. 

The total loss function is a weighted sum  of a Mean Average Error (MAE) , referred as $\mathcal{L}_{1}$ and used in ~\cite{NVS_skip}), and a second term, called $\mathcal{L}_{spectral}$, directly inspired form prior super-resolution work ~\cite{SR} and that extensively focuses on the preserving high frequencies. Indeed, considering a 2D Gaussian filter $w_{gauss}$, an image $I$ can be decomposed into a low and a high-frequency components, respectively noted as $I^{LF}$ and $I^{HF}$ : 
\begin{equation}
\begin{cases}
     I^{LF}  = I\circledast w_{gauss} \\
     I^{HF} = I - I^{LF} = (\delta - w_{gauss})\circledast I
\end{cases}
\end{equation}
where $\circledast$ represents a 2D convolution operation. 
The final loss function used during training is thus given by: 
\begin{equation}
    \mathcal{L} = \mathcal{L}_{1} + \lambda \mathcal{L}_{spectral} = |\hat{I}_{t} - I_{t}| + \lambda \left( \hat{I}_{t}^{HF} - I_{t}^{HF} \right)^{2}
    \label{eq:1}
\end{equation}

Additional details on $\mathcal{L}_{spectral}$ are available in the Supplementary.

\section{Experiments}
All qualitative and quantitative results we obtained through our camera pose encoding strategy are presented in this section. \newline

\noindent\textbf{Datasets.} We experimented our method on the same dataset as ~\cite{NVS_skip, sun2018multiview}: on the \textit{chair} and \textit{car} class from ShapeNet ~\cite{Shapenet} as well as on real scene images from KITTI ~\cite{KITTI} and Synthia ~\cite{Synthia}.
Results we reported for ~\cite{NVS_skip} slightly differ from the ones published since we get consideration for a more challenging rendering scenario: jittered camera pose in ShapeNet ~\cite{Shapenet}, elements on image borders from Synthia ~\cite{Synthia} and KITTI ~\cite{KITTI} are not removed anymore through  center-cropping , etc. Such changes are motivated and fully exposed in the Supplementary.  \newline

\noindent\textbf{Training - Testing.} We trained our model on 100,000 iterations, using the same training procedure as ~\cite{NVS_skip}. Since our camera pose encoding strategy is light enough, we perform ``on the fly'' batch creation during training. Inference scores were averaged over 100 runs with batch sizes of 16. \newline

\noindent\textbf{Metrics.} We used the Mean Average Error (MAE),  Structural Similarity Index Measure (SSIM) ~\cite{SSIM} and Peak Signal-to-Noise Ratio (PSNR) metrics to score our novel view synthesis architecture. 

\subsection{Qualitative Results}

\begin{figure*}[h!]
    \begin{center}
    \includegraphics[width=.8\textwidth]{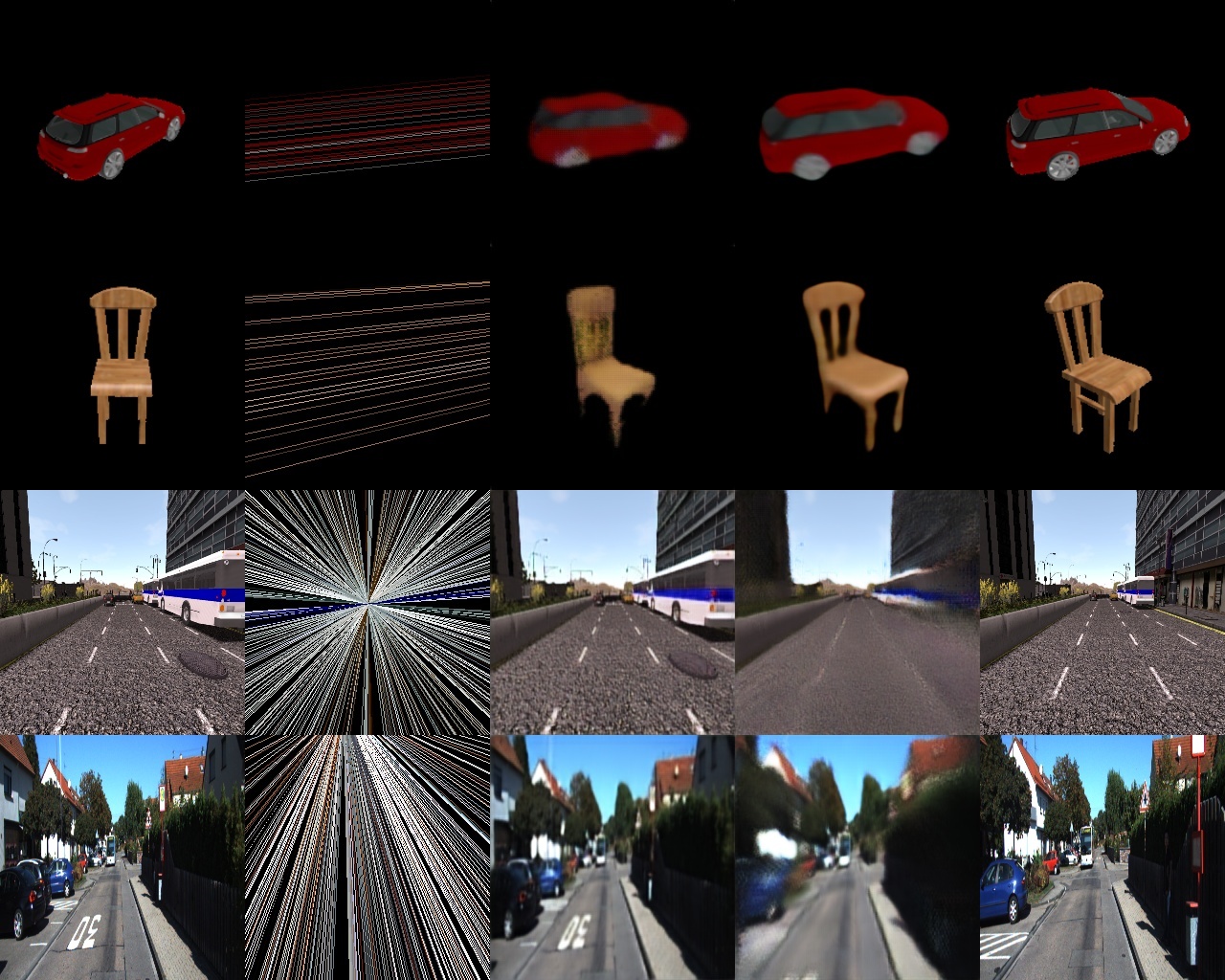}
    \end{center}
     \caption{Inference results on all the four datasets considered. $E_{s\xrightarrow{}t}$ encoded transformation have been obtained with $\textbf{G}_{15}$. From left to right: the source image  $I_s$, $E_{s\xrightarrow{}t}$, ~\cite{NVS_skip} prediction, our prediction, the target image $I_t$. On overall, our method predicts more consistent results since complete pose transformation has been fed to the network contrary to the concurrent work from ~\cite{NVS_skip}. }
     \label{fig:res_all}
\end{figure*}

We present in Figure \ref{fig:res_all} an instance of all the datasets we get consideration for. Whereas our model manages to infer and reason about the object size according to the target view on the \textit{Car} and \textit{Chair} classes, novel-view produced by ~\cite{NVS_skip} fails to do so, producing a result that rather roughly matches the source object size. Our results on these two classes are also sharper and more realistic, both from colour and geometrical perspectives. The car's wheels or the chair's complex back are thus better synthesise with our method. 

Because ~\cite{NVS_skip} integrates the camera pose as discretized bins, the viewpoint transformation loses its physical inner 3D consistency structure. On the other hand, our method produces an encoding that fully accounts for the continuous pose transformation that occurred.\newline

Results of our method synthesised on real-world datasets are depicted on third (Synthia ~\cite{Synthia}) and fourth (KITTI ~\cite{KITTI}) rows of the Figure \ref{fig:res_all}. One might noticed how the car moved between the source and the target view in the Synthia ~\cite{Synthia} dataset. While our results remain blurry, our network architecture successfully hallucinated the forward motion that the bus had. On the other hand, our main concurrent work mostly predicts the source image, and fails to capture the relative displacement that occurred in this example. An almost similar scenario is inspected on KITTI ~\cite{KITTI}. While the sign "30" on the ground has disappeared between the two views, our method successfully captures such drastic change while ~\cite{NVS_skip} fails to the same extent as previously. We emphasise the fundamental role our camera encoding strategy has here, especially regarding the shape of the various objects where the perspective changes a lot between $I_s$ and $I_t$ (on the roof of the house, top left border). \newline

\begin{figure*}[h!]
    \begin{center}
    \includegraphics[width=.8\textwidth]{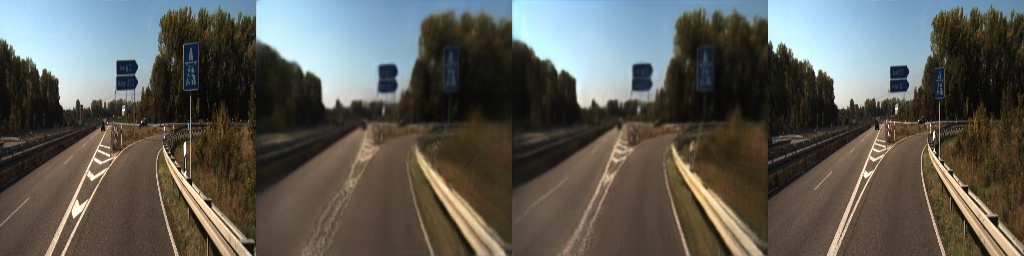}
    \end{center}
     \caption{From left to right: the source image  $I_s$, ~\cite{NVS_skip} prediction, our prediction, the target image $I_t$. Painting on the ground are better reconstructed in our method compared to ~\cite{NVS_skip}.}
     \label{fig:res_all2}
\end{figure*}

While results presented on Figure \ref{fig:res_all} for real world datasets focus on the overall motion that occurred, we highlight on Figure \ref{fig:res_all2} in which extent high-frequency details can be retrieved by our model. The global motion is quite properly handled by ~\cite{NVS_skip} too, but our method manages to better retrieved tiny details structure for the paintings on the ground. The Spectral loss we used during training helped the network to handle these high frequencies. A complete ablation study regarding the impact the Spectral loss has can be found in the Supplementary, in addition to several other visual results on the four datasets we considered.

\subsection{Quantitative Results }

We report some performance results over the four datasets we get consideration for. Tables \ref{tab:1} and \ref{tab:2} respectively summarise the scores for the synthetic ShapeNet ~\cite{Shapenet} dataset and the real-scene ones (Synthia ~\cite{Synthia} and KITTI ~\cite{KITTI}). While our method outperforms concurrent works on MAE, we reach extremely competitive results with the SSIM metric too, since only ~\cite{sun2018multiview} gets better scores. \newline

However, we emphasise on a crucial aspect regarding the reported scores. Only our method and the one from ~\cite{NVS_skip} has been retrained with the extended and challenging datasets that we presented earlier. Results from remaining works ~\cite{sun2018multiview,22,23,19} are the ones that were originally published by authors. 
\begin{table}[htp!]
\begin{center}
\begin{adjustbox}{max width=\textwidth}
\begin{tabular}[h]{c||ccccccc}
\hline
Modality & Method & \multicolumn{3}{c}{Car} & \multicolumn{3}{c}{Chair} \\
 & & $MAE$ ($\downarrow$) & SSIM ($\uparrow$) & PSNR ($\uparrow$) & $MAE$ ($\downarrow$) & SSIM ($\uparrow$)) & PSNR ($\uparrow$) \\
\hline
\textit{Multi-views} &~\cite{sun2018multiview} & $0.078$ & ${0.935}$ & - & $0.141$ & ${0.911}$ & -\\
\hline
& ~\cite{19} & $0.139$ & $0.875$ & - & $0.223$ & $0.882$ & -\\
&~\cite{22} & $0.148$ & $0.877$ & - & $0.229$ & $0.871$ & - \\
\textit{Single-view} &~\cite{23} & $0.119$ & $\underline{0.913}$ & - & $0.202$ & 0.889& -\\
& ~\cite{pixelnerf} & - & 0.900 & \underline{23.17} & - & $\mathbf{0.911}$ & $\mathbf{23.72}$ \\
&~\cite{NVS_skip} & $\underline{0.026}$ & $0.892$ & 21.18 & $\underline{0.045}$ & $0.865$ & 17.89\\
& Ours & $\mathbf{0.016}$ & $\mathbf{0.928}$ & $\mathbf{24.23}$ & $\mathbf{0.032}$ & \underline{0.901}& \underline{19.55}\\
\hline \hline

\end{tabular}
\end{adjustbox}
\end{center}
\captionof{table}{Performance on ShapeNet ~\cite{Shapenet}. Best scores for the single-view modality are highlighted in bold while second best ones are underlined.}
\label{tab:1}
\end{table}

 Results on Synthia~\cite{Synthia} and KITTI ~\cite{KITTI} are reported on Table \ref{tab:2} and are competitive with current state of the art methods. We emphasize that our method (as well as ~\cite{NVS_skip}) was trained on more complex scenarios from an image-content perspective: both Synthia and KITTI images were resized to $256\times 256$ without any center-cropping operation (which allows to discard the most challenging elements from the scenes). 

\begin{table}[htp!]
\begin{center}
\begin{adjustbox}{max width=\textwidth}
\begin{tabular}[h]{c||ccccccc}
\hline
Modality &Method & \multicolumn{3}{c}{Synthia} & \multicolumn{3}{c}{KITTI} \\
& & $MAE$ ($\downarrow$) & SSIM ($\uparrow$)& PNSR ($\uparrow$) & $MAE$ ($\downarrow$) & SSIM ($\uparrow$) & PNSR ($\uparrow$)\\
\hline
\textit{Multi-views}&~\cite{sun2018multiview} & $0.118$ & $0.737$ & - &  $0.163$ & $0.691$ & - \\
\hline
&~\cite{19} & $0.175$ & $0.612$ & - &  $0.295$ & $0.505$ &  - \\
\textit{Single-view} & ~\cite{22} & $0.221$ & $\mathbf{0.636}$ & -  & $0.418$ & $0.504$ & - \\
 & ~\cite{NVS_skip}  & $\underline{0.065}$ & $\underline{0.632}$ & $\mathbf{19.81}$ & $\underline{0.087}$ & $\underline{0.602}$ & \underline{16.84} \\
& Ours & $\mathbf{0.065}$ & $0.631$ & \underline{19.44} & $\mathbf{0.082}$ & $\mathbf{0.609}$ & $\mathbf{17.11}$ \\
\hline\hline
\end{tabular}
\end{adjustbox}
\end{center}
\captionof{table}{Performance on Synthia ~\cite{Synthia} and KITTI ~\cite{KITTI}. Best scores for the single-view modality are highlighted in bold while second best ones are underlined.}
\label{tab:2}
\end{table}

\section{Limitations and further works}
The pose encoding strategy introduced in this paper is an innovative and elegant solution to integrate the camera pose transformation in the novel view synthesis issue. However, our method suffers from some limitations that might be tackled to reach even better results, both from image quality and timely processing perspectives. Indeed, since we compute the encoded viewpoint transformation $E_{s\xrightarrow{}t}$ "on the fly" during training, our method is slower than our main concurrent work ~\citep{NVS_skip}. Another further line of work concerns the camera data requirements that could be more flexible since one might argue that our method requires intrinsic parameters in addition to the extrinsic ones. 

One might finally also think for future work about leveraging onto the encoded relative poses we designed to better constraints the network during training through a cyclic loss function. Indeed, it would be possible to define a reversed encoded relative pose through $E_{t\xrightarrow{}s}$, by leveraging the predicted novel view.

\section{Conclusion}

In this paper, we proposed a new and innovative method to encode the camera transformation for the deep learning-based novel view synthesis task. For this, we leverage epipolar geometry in order to encode such viewpoint displacement as an image that we call the encoded relative pose $E_{s\xrightarrow{}t}$ (made of several coloured epipolar lines). We argue this new camera transformation encoding is better suited for the single-image novel view synthesis issue than the standard way that only consists in considering the extrinsic values of the camera transformation. Indeed, the idea behind the vanilla approach is to feed the neural network with an RGB source image and a camera viewpoint transformation to generate a new image that can be viewed as the impact of such displacement on the input image. Our method rather proposes to provide both an RGB source image and the encoded image of the viewpoint transformation, which is already a strong insight into the impact of the displacement on this image. In other words, the core motivation of our proposed method is to help the network to better understand the correlation between the input image and the desired camera viewpoint transformation. The experimental results on the different datasets presented in this work confirm our claim. These results tend to prove that this new encoding strategy is more robust to complex displacements, namely large perspective changes and also generates images with sharper details. 

%%%%%%%%%%%%%%%%%%%%%%%%%%%%%%%%%%%%%
%%%%%%% SUPPLEMENTARY MATERIALS%%%%%%
%%%%%%%%%%%%%%%%%%%%%%%%%%%%%%%%%%%%%
\bibliography{egbib}
\newpage
\appendix

\section{Extended encoding strategy}
\label{sec:extended}
As explained in the core paper, we extended our original relative pose encoding strategy to better apprehend the specificity of KITTI ~\cite{KITTI} and Synthia ~\cite{Synthia} datasets. 

\begin{figure}[ht]
\vskip 0.2in
\centering
\begin{subfigure}
  \centering
  \includegraphics[width=.4\linewidth]{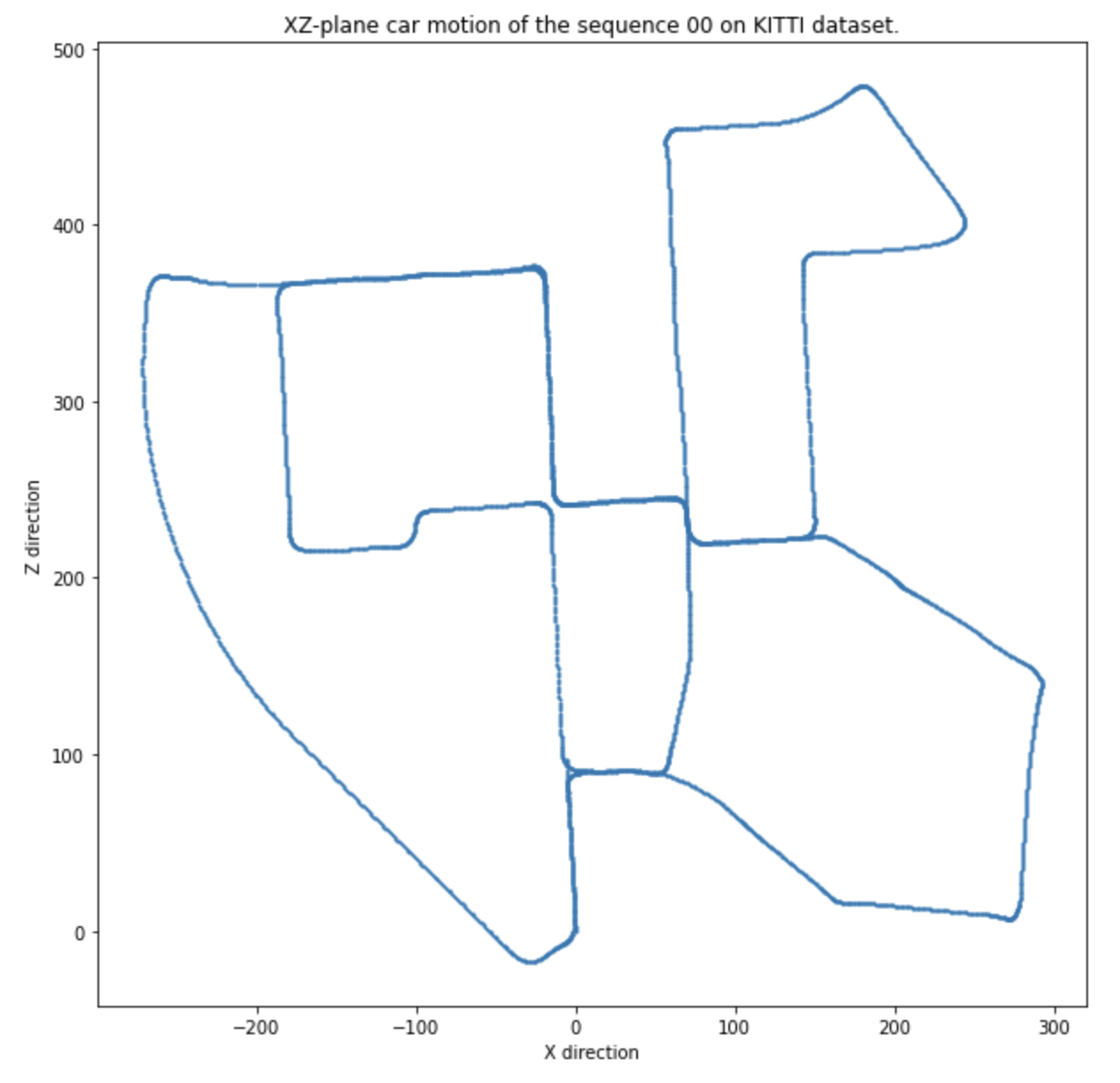}
\end{subfigure}%
\begin{subfigure}
  \centering
  \includegraphics[width=.4\linewidth]{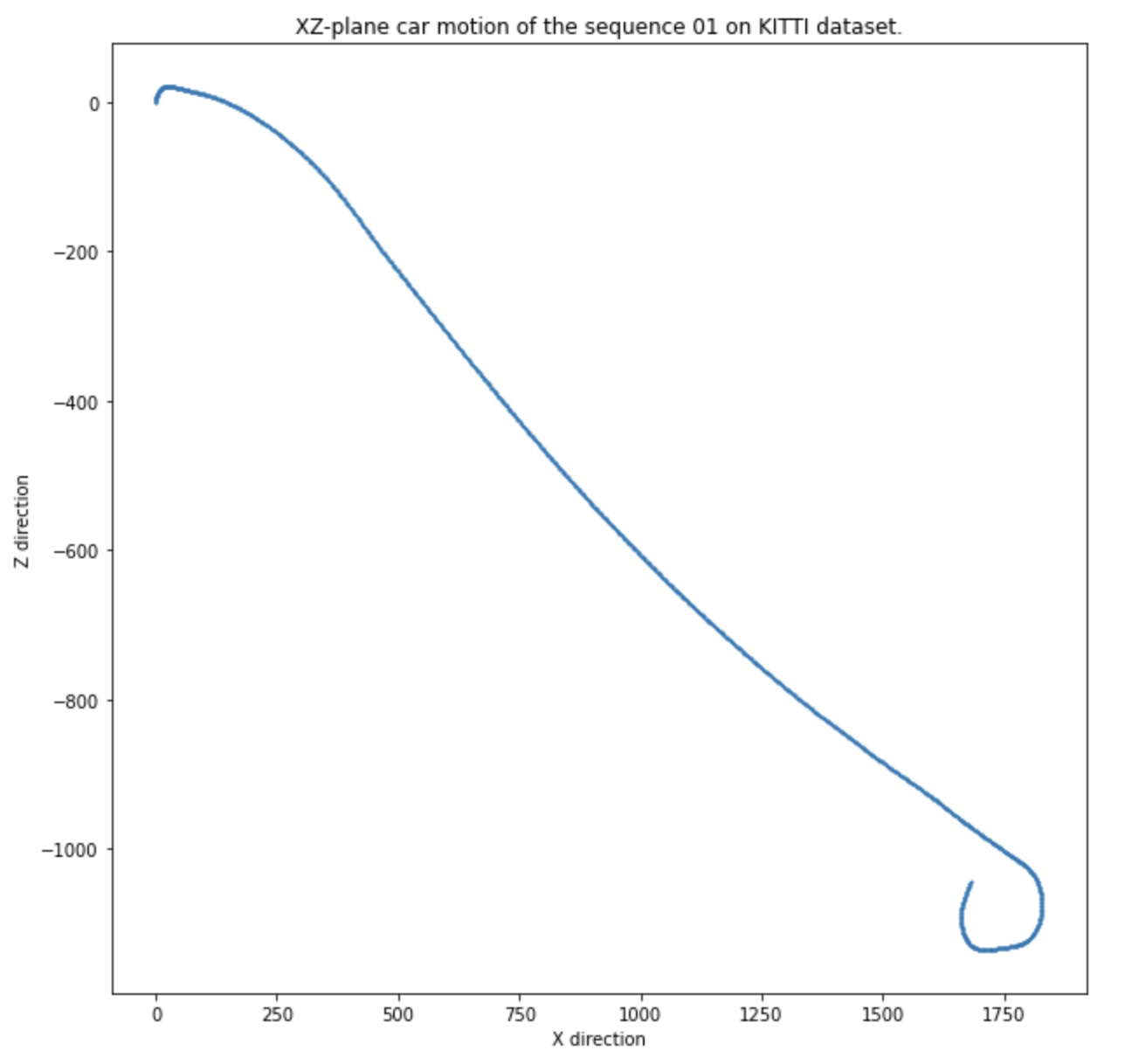}
\end{subfigure}
\caption{We illustrate below the overall trajectory in the (XZ) plane of the car that drove across German's street to acquire the 00 (left) and 01 (right) sequence of the KITTI~\cite{KITTI} dataset.}
\label{fig:test}
\vskip -0.2in
\end{figure}

As shown on the Figure \ref{fig:test}, most of the car trajectories are made based on a straight path and only a few turns exist in the sequence \textbf{00}. In a similar fashion way, the sequence \textbf{01} is almost an end-to-end complete pure translation.

\section{Spectral loss} 

Theoretical insights are presented in this section regarding the spectral loss function that was used for training in addition to the usual MAE. 

As already mentioned in the main paper, such loss directly takes inspiration from ~\cite{SR}, one of the latest state-of-the-art paper related to the super-resolution issue. Authors emphasise in their work on the fundamental role high frequencies have in the image generation process. \newline

The Gaussian filter $w_{gauss}$ we used is straightforwardly defined by a mean $\mu = \frac{k_{s}-1}{2}$ and a variance $\sigma = (\frac{k_{s}}{k_{s}+1})^{2}$, with $k_{s}$=5.

Such formulation allows to end up with the  Spectral loss function that was introduced in the paper: 
\begin{equation}
    \mathcal{L}_{spectral} = ||I_{t}^{HF} - \hat{I}_{t}^{HF}||_{2}^{2} 
\end{equation}
We therefore extensively focus through this loss on the highest frequencies of the target image, to enforce during training the network to retrieve as much as much as possible fine and complex structures. 

\section{Experiments}
\subsection{Dataset characteristics}
We first provide here some additional information regarding the different datasets we used for our experiments. 

Regarding the ShapeNet dataset, we decided to not use the same dataset as ~\cite{NVS_skip, sun2018multiview}  but rather worked with the rendered ShapeNet images from DISN ~\cite{DISN}. It offers at least three main improvements over the ones used in ~\cite{NVS_skip}:

\begin{itemize}
    \item Intrinsic camera parameters are available. 
    \item Each object within a class has 36 different views (against 18 for the dataset provided by ~\cite{NVS_skip}). 
    \item Rendered images have a non null elevation angle and the azimuth one is sampled on a regular 10°  basis. A random noise term is added on each rendered view to slightly jitter the camera pose. 
\end{itemize}

Considering real-world Synthia ~\cite{Synthia} and KITTI ~\cite{KITTI} datasets, original images used in ~\cite{NVS_skip} also only contain extrinsic matrices, leaving apart the intrinsic information our architecture requires. We, therefore, build up our own train/test sets, with the same scenes as the ones used in ~\cite{NVS_skip}. Images were resized to $256\times256$ for speed-up and convenience purposes and ground-truth intrinsic matrices were adjusted accordingly. Images we work with on these real scene scenarios are more challenging than the ones used in ~\cite{NVS_skip}~\cite{sun2018multiview} since dealing with center-cropped images discards fast-moving elements (on image borders) from the scenes.

Finally, we get consideration for the same setting used in ~\cite{NVS_skip} for the maximal latitude between the source and target view: full $\pm 180^{\circ}$ azimuthal range is permitted for the ShapeNet classes while a maximum of $\pm 10$ frames are considered for the real world datasets Synthia~\cite{Synthia} and KITTI ~\cite{KITTI}.

\subsection{Ablation studies}
\label{sec:ablation}
We conduct ablation studies to highlight and understand how some meaningful properties of our encoding strategy behave. \newline

\textbf{Benefit of the extended encoded pose strategy}

Dealing with real-world scene datasets and the maximum 10 frames difference that could occurred between the source and the target view is one of the most tricky scenarios for single-image novel view synthesis. We thus conduct a first ablation study to validate the intuition behind this additional channel that encodes the relative largest motion in the (XZ) plane. Please note that the Spectral loss has not been added in this ablation study, leading to slightly different results from the original ones reported in the main paper. 

\begin{table}[h!]
\begin{center}
\begin{tabular}{ccccc}
\hline
Method & \multicolumn{2}{c}{Synthia} & \multicolumn{2}{c}{KITTI} \\
 & $MAE$ ($\downarrow$) & SSIM ($\uparrow$) & $MAE$ ($\downarrow$) & SSIM ($\uparrow$) \\
\hline
Encoded pose &  0.077 & 0.602  & 0.109  & 0.576 \\ \hline
Extended encoded pose &  \textbf{0.066} &  \textbf{0.622} &  \textbf{0.086} &  \textbf{0.605}  \\ \hline
\hline
\end{tabular}
\end{center}
\caption{Benefit on Synthia ~\cite{Synthia} and KITTI ~\cite{KITTI} datasets of our extended encoding strategy. Adding such fourth channel helps the network to better perform on real-world datasets. The grid $\textbf{G}_{15}$ has been used in both cases.}
\label{tab:compExtended}
\end{table}

As shown in Table \ref{tab:compExtended}, and considering the neural network architecture as fixed, the fourth channel we have added to our representation $E_{s\xrightarrow{}t}$ clearly helps the network to perform better in the task it has been trained for.  The SSIM gained more than 2 points on average while the MAE significantly decreased (by almost 20\% on average for the two datasets) to reach competitive results with ~\cite{NVS_skip}. 

\begin{figure}[htp]

\begin{center}
\includegraphics[width=.45\textwidth]{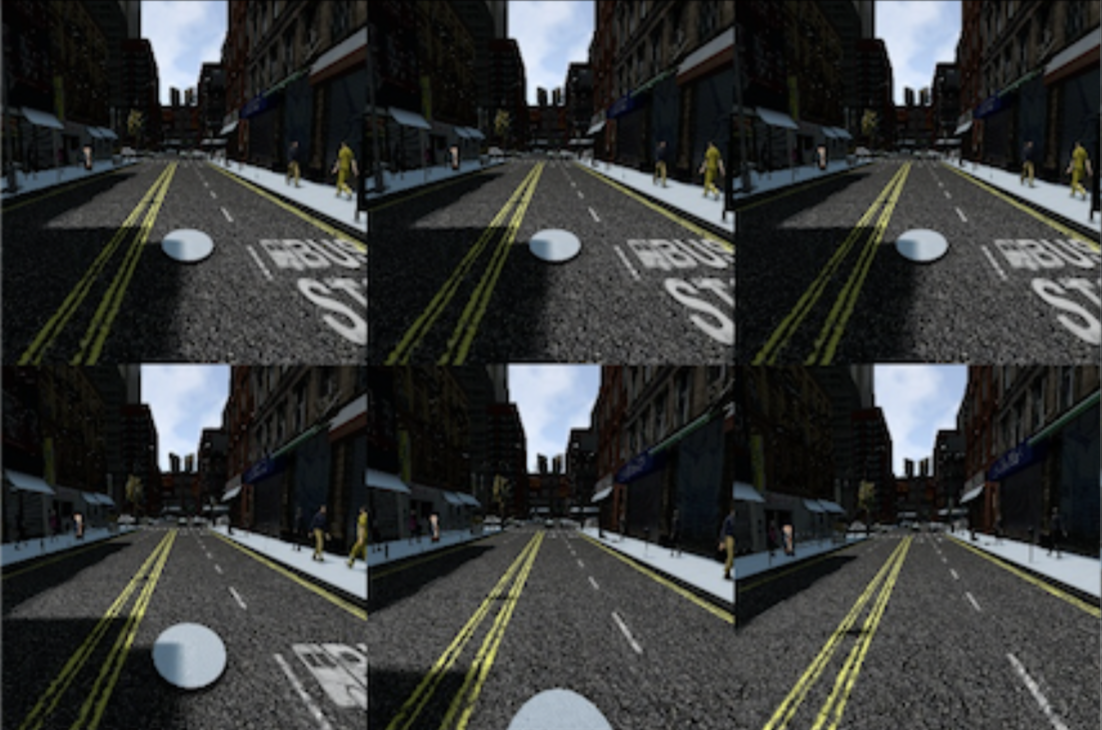} \hfill
\includegraphics[width=.45\textwidth]{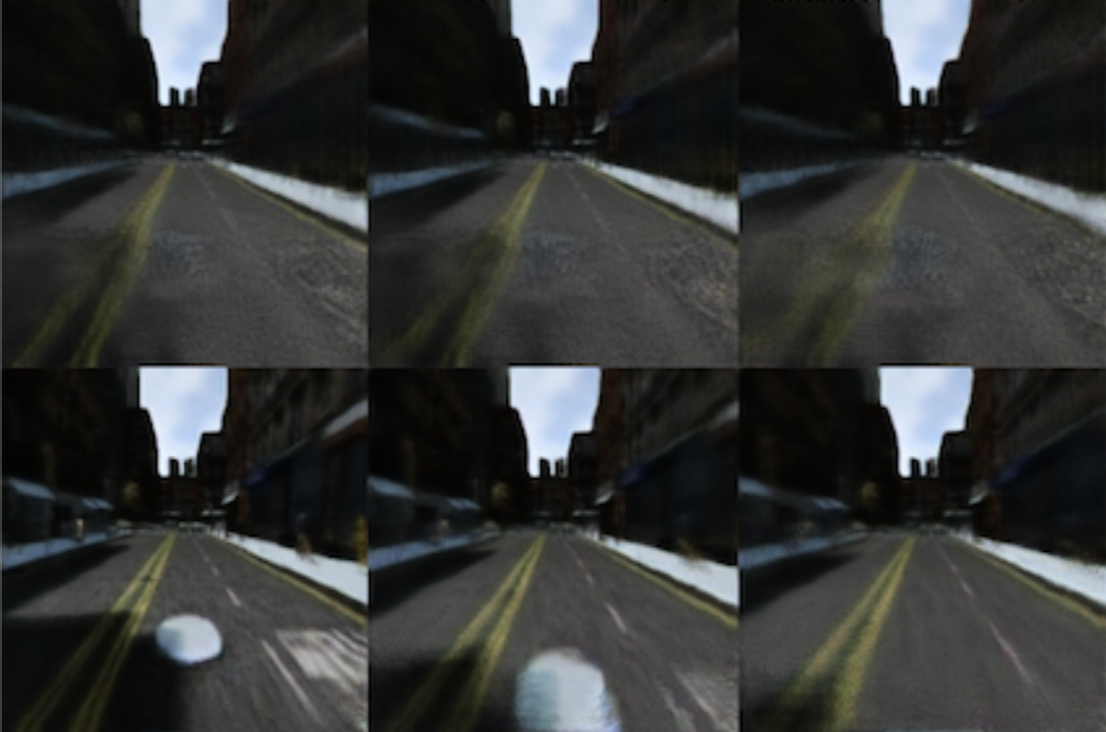}
\end{center}
\caption{Source fixed view $I_s$ (top row, Left) and three consecutive target views (bottom row, Left) from Synthia ~\cite{Synthia} test set. - Predictions made by our model with the encoded pose (top row,Right) and the extended encoded pose strategy (bottom row,Right). Adding an additional channel in our extended pose encoding allows the network to better apprehend the motion that occurred in Synthia ~\cite{Synthia} and KITTI~\cite{KITTI}. The grid $\textbf{G}_{15}$ has been used in both cases.}
\label{fig:ablaSynthia}
\end{figure}
We highlight on Figure \ref{fig:ablaSynthia} the positive influence this last channel has on the our model. 

While the manhole cover (disappearing on the target views sequence) is entirely discarded by the network trained with the 3 channel pose encoding representation, the extended version we proposed managed to grasp the car's motion.\newline 

\textbf{Spectral loss influence}

A second ablation study has been conducted to highlight to which extent the spectral loss function positively impacts the training of our model architecture. 
\begin{table}[h!]
\begin{center}
\begin{tabular}{@{}lllllllllllllllll@{}}
  \toprule
  Datasets & Metrics  &$\mathcal{L}_{1}$ only  &  $\mathcal{L}_{1}+\mathcal{L}_{spectral}$ &   \\
  \midrule
  \multirow{3}{*}{\textbf{ShapeNet - Car}} & L1 ($\downarrow$) &\hfil 0.019 & \hfil\textbf{0.016} \\
  & SSIM ($\uparrow$) & \hfil0.912 & \hfil\textbf{0.928}\\
  & PSNR ($\uparrow$)& \hfil22.61 & \hfil\textbf{24.23}\\
  \midrule
  \multirow{3}{*}{\textbf{ShapeNet - Chair}} & L1 ($\downarrow$) & \hfil 0.037 & \hfil \textbf{0.032}\\
  & SSIM ($\uparrow$)&\hfil 0.892 & \hfil \textbf{0.901} \\
  & PSNR ($\uparrow$) & \hfil 19.19 & \hfil \textbf{19.55} \\
  \midrule
  \multirow{3}{*}{\textbf{Synthia}} & L1 ($\downarrow$)& \hfil 0.066 & \hfil \textbf{0.065}\\
  & SSIM ($\uparrow$)& \hfil 0.622 & \hfil \textbf{0.631} \\
  & PSNR ($\uparrow$)& \hfil 19.24 & \hfil \textbf{19.44}\\
  \midrule
  \multirow{3}{*}{\textbf{KITTI}} & L1 ($\downarrow$)& \hfil 0.086 & \hfil \textbf{0.082}\\
  & SSIM ($\uparrow$)& \hfil 0.605 & \hfil \textbf{0.609} \\
  & PSNR ($\uparrow$)& \hfil 16.99 & \hfil \textbf{17.11}\\\hline
\hline
\end{tabular}
\end{center}
\caption{Impact of the Spectral loss function.}
\label{tab:spectral}
\end{table}

\begin{figure*}[h!]
    \begin{center}
    \includegraphics[width=.5\textwidth]{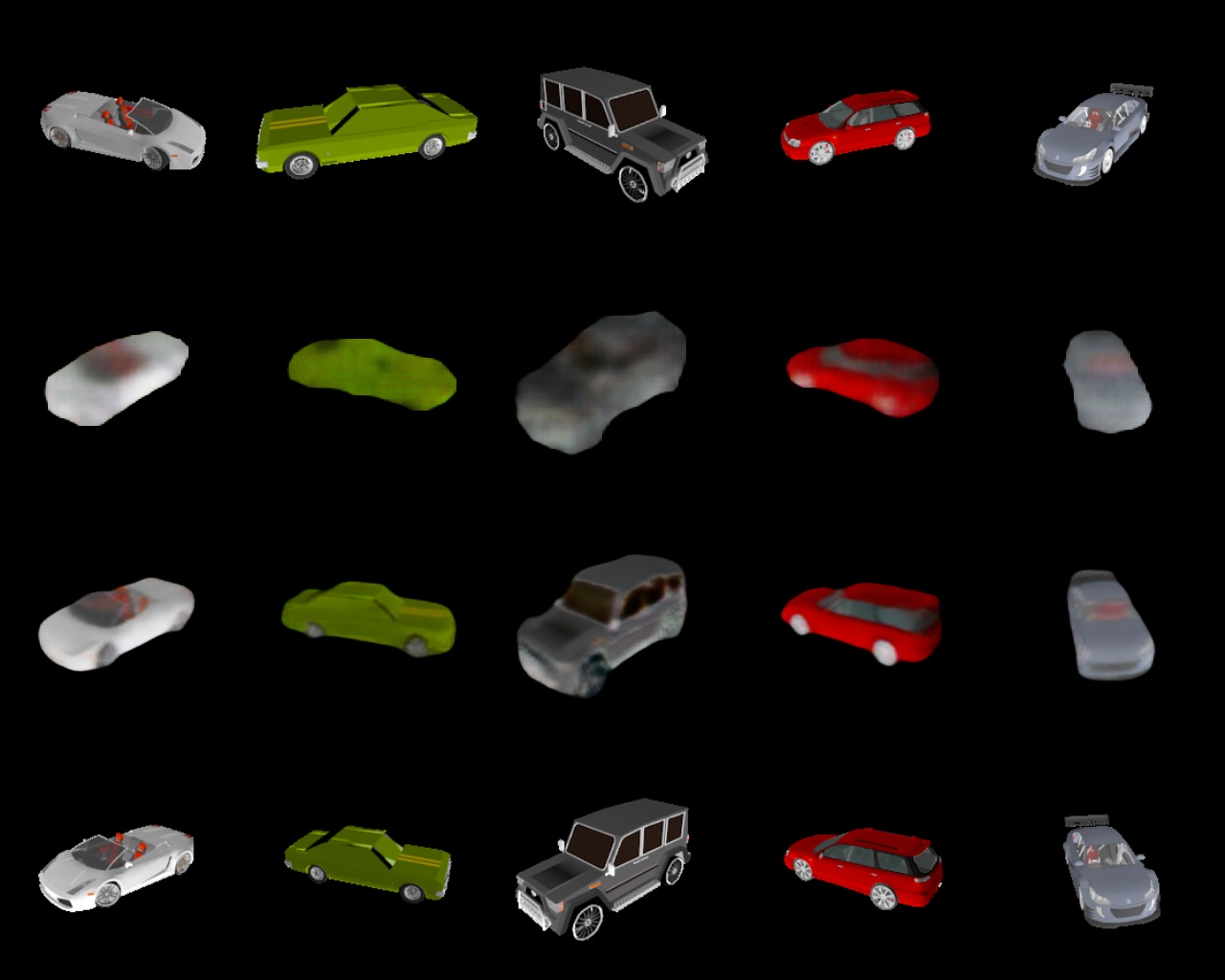}
    \end{center}
     \caption{Inference results from the ShapeNet ~\cite{Shapenet} \textit{Car} test set. From the top row to the bottom one: Source images  $I_s$, Ours prediction with $\mathcal{L}_{1}$ only at training, Ours prediction with  $\mathcal{L}_{spectral} + \mathcal{L}_{1}$ at training, Ground truth - Target images $I_t$. From a general perspective, tires and windows are better retrieved at inference time when high frequencies have been constrained during training.}
     \label{fig:spectral_res}
\end{figure*}

As seen in Table \ref{tab:spectral}, constraining the training on the high frequencies helps the network to generate more realistic novel views. Such quantitative improvement is visually confirmed in Figure \ref{fig:spectral_res} where the same object instance is generated through both configurations with the ShapeNet \textit{Car} class.\newline 

\newpage\textbf{Discrete grid $\textbf{G}_{r}$ granularity and sampling strategy}

We present a third ablation study to get some knowledge regarding the granularity our grid sampling needs. Beyond the three grids we tested out, we also consider a random sampling strategy, that consists of sampling a fixed number (corresponding to 1\% of pixels for a $256\times 256$ image) of locations. Table \ref{tab:3} summarises the different results of this experiment on the real-world datasets.
\begin{table}[h!]
\begin{center}
\begin{tabular}{ccccc}
\hline
Method & \multicolumn{2}{c}{Synthia} & \multicolumn{2}{c}{KITTI} \\
 & $MAE$ ($\downarrow$) & SSIM ($\uparrow$) & $MAE$ ($\downarrow$) & SSIM ($\uparrow$) \\
\hline
\makecell{ Random Sampling \\ \footnotesize{(655 pix. sampled)}}& \textbf{0.0816} & \textbf{0.593} & 0.1290 & 0.549   \\ \hline
\makecell{$\textbf{G}_{15}$ grid \\ \footnotesize{(225 pix. sampled)}} & 0.0823 & 0.589 & 0.1222 & 0.562   \\ \hline
\makecell{$\textbf{G}_{20}$ grid \\ \footnotesize{(400 pix. sampled)}} & 0.0857 & 0.576 & 0.1241 & 0.560  \\ \hline
\makecell{$\textbf{G}_{25}$ grid \\ \footnotesize{(625 pix. sampled)}} & 0.0908 & 0.575 & \textbf{0.1217} & \textbf{0.563}   \\ \hline
\hline
\end{tabular}
\end{center}
\caption{Sampling strategy influence over the real-world datasets Synthia ~\cite{Synthia} and KITTI ~\cite{KITTI}.}
\label{tab:3}
\end{table}

Overall, there are no significant differences between the strategies that were tested in this ablation study.  However, the random and the grid sampling strategy differ in an important aspect: the latter performs significantly faster and roughly takes 4 times less time to form a batch of triplets $\left(I_{s},I_{t},E_{s\xrightarrow{}t}\right)$ than the random sampling strategy. \newline

Using a regular grid $\textbf{G}_{r}$ always use the same pixel locations from $I_s$ to build  $E_{s\xrightarrow{}t}$ while the random sampling strategy imposes to picked up new samples all the time.  

We performed the same ablation study on the synthetic ShapeNet dataset ~\cite{Shapenet} and drew identical observations. 

\subsection{Inference results}

We finally present in this last part additional qualitative results from our model, on all the four datasets we have get consideration for in this study. We present for each dataset 8 different scenes or object instances. \newline

\begin{figure*}[htp!]
    \begin{center}
    \includegraphics[width=.9\textwidth]{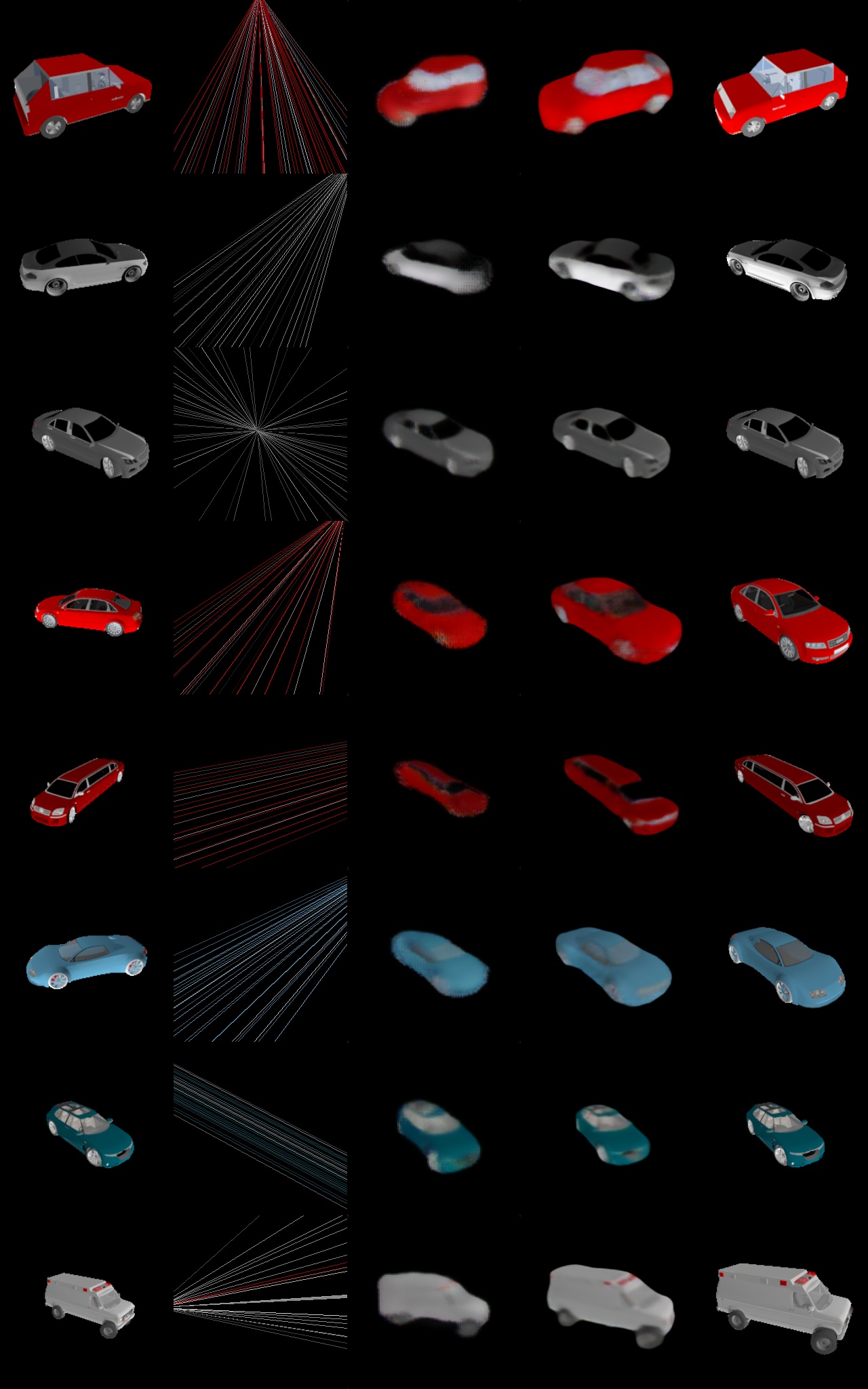}
    \end{center}
     \caption{Visual results from the test ShapeNet ~\cite{Shapenet} \textit{Car} class. From left to right: the source image  $I_s$, the Encoded Pose $E_{s\xrightarrow{}t}$,  the prediction of ~\cite{NVS_skip}, our prediction and the target image $I_t$.}     \label{fig:add_visCar}
\end{figure*}

\begin{figure*}[htp!]
    \begin{center}
    \includegraphics[width=.9\textwidth]{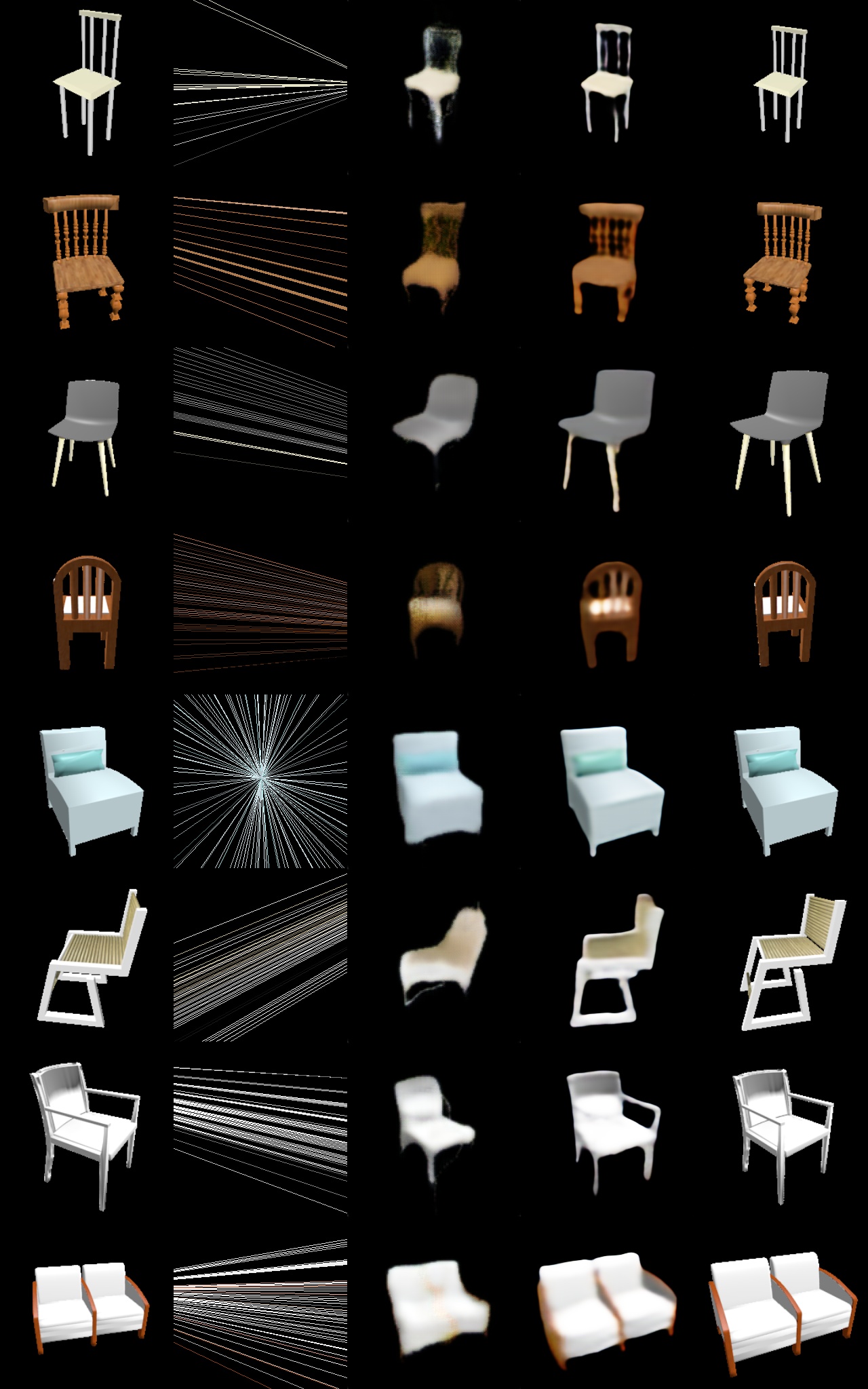}
    \end{center}
     \caption{Visual results from the test ShapeNet ~\cite{Shapenet} \textit{Chair} class. From left to right: the source image  $I_s$, the Encoded Pose $E_{s\xrightarrow{}t}$,  the prediction of ~\cite{NVS_skip}, our prediction and the target image $I_t$.}
     \label{fig:add_visChair}
\end{figure*}

\begin{figure*}[htp!]
    \begin{center}
    \includegraphics[width=.9\textwidth]{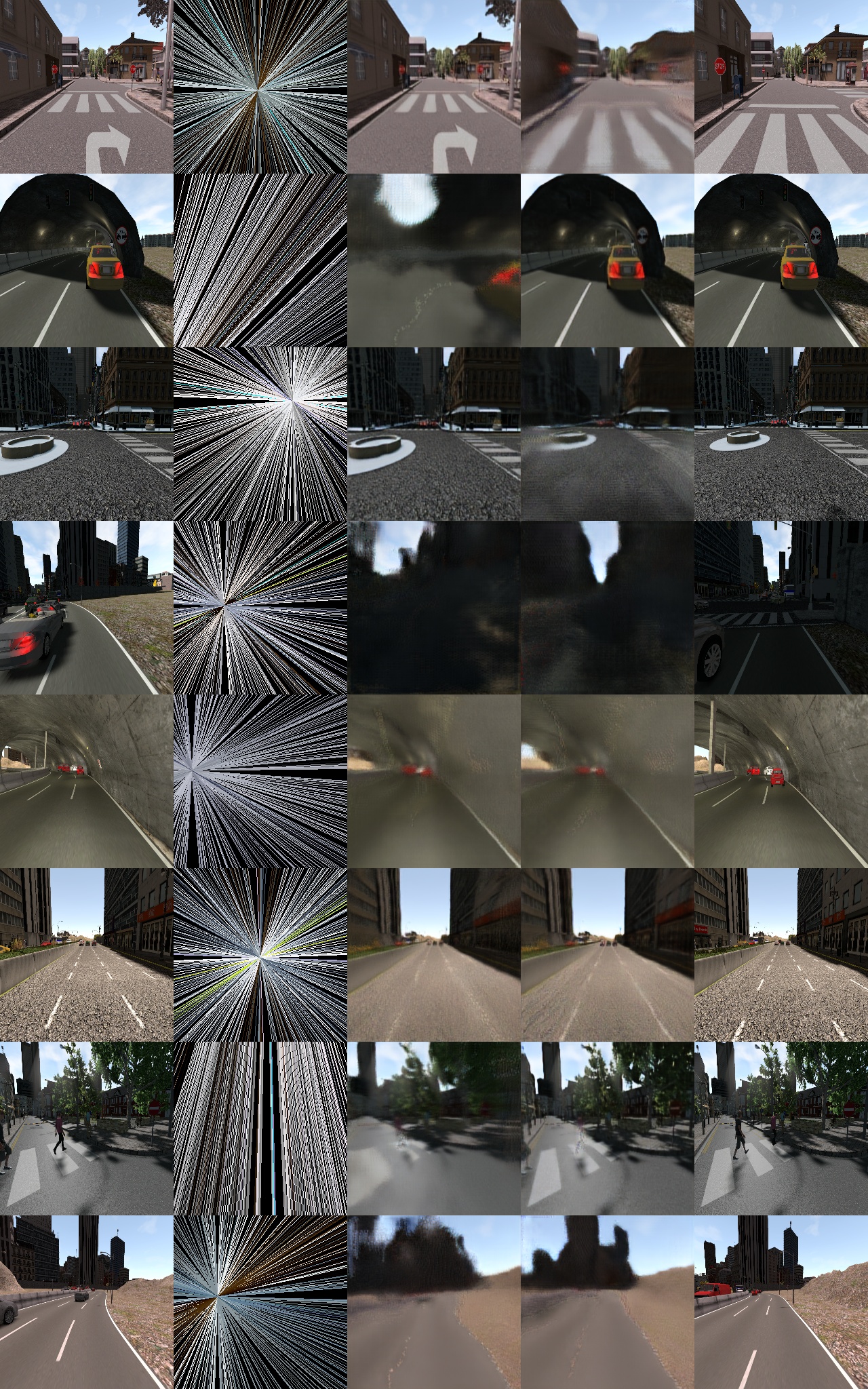}
    \end{center}
     \caption{Visual results from the Synthia ~\cite{Synthia} test set. From left to right: the source image  $I_s$, the Encoded Pose $E_{s\xrightarrow{}t}$,  the prediction of ~\cite{NVS_skip}, our prediction and the target image $I_t$.}
     \label{fig:add_visSynthia}
\end{figure*}

\begin{figure*}[htp!]
    \begin{center}
    \includegraphics[width=.9\textwidth]{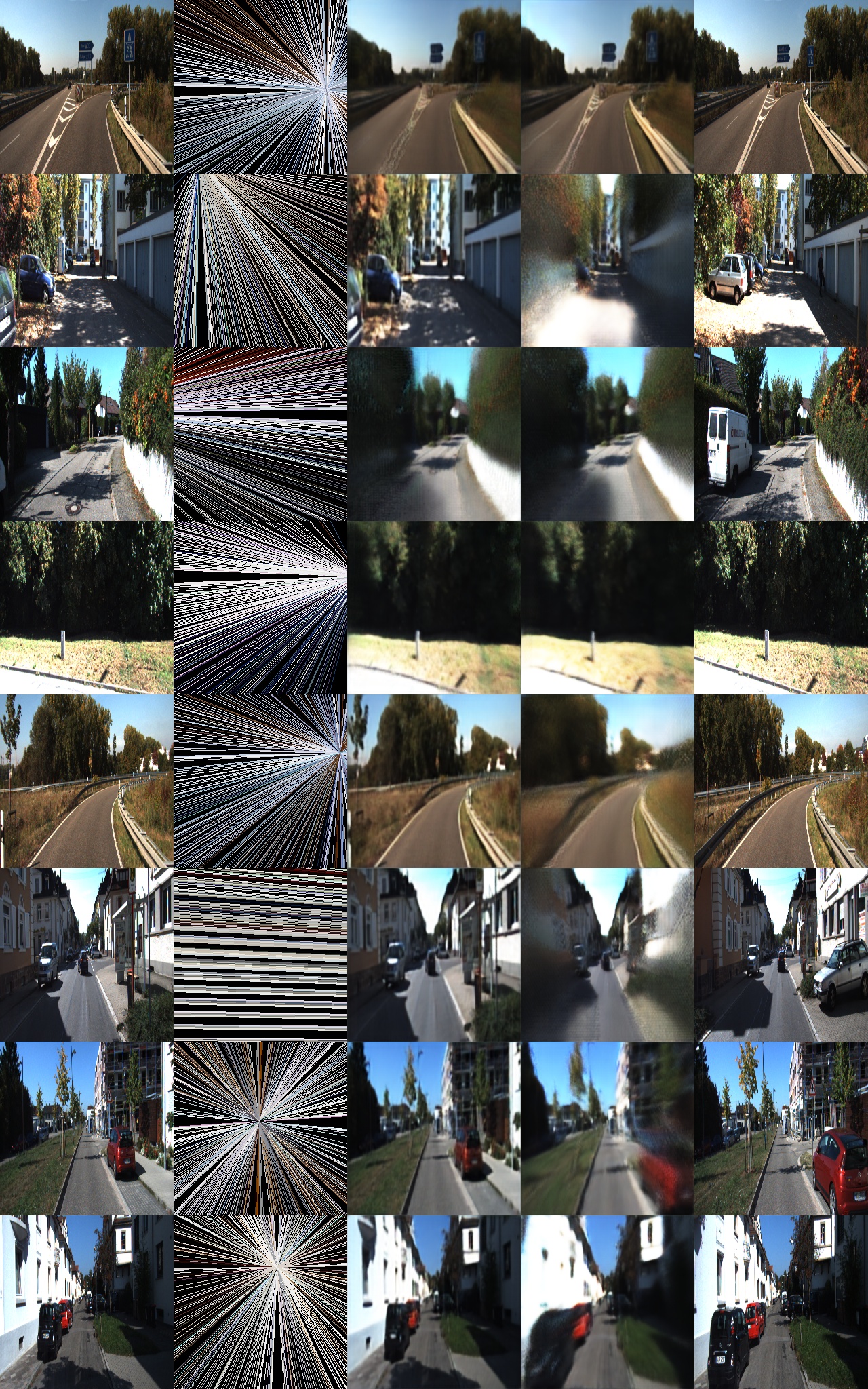}
    \end{center}
     \caption{Visual results from the KITTI ~\cite{KITTI} test set. From left to right: the source image  $I_s$, the Encoded Pose $E_{s\xrightarrow{}t}$,  the prediction of ~\cite{NVS_skip}, our prediction and the target image $I_t$.}
     \label{fig:add_visKITTI}
\end{figure*}
\clearpage 
\end{document}